# Assessment of central serous chorioretinopathy (CSC) depicted on color fundus photographs using deep Learning

Yi Zhen[1], Hang Chen[2], Xu Zhang[1], Meng Liu[1], Xin Meng[2], Jian Zhang[2], Jiantao Pu[2]

[1] National Engineering Research Center for Ophthalmology, Beijing Institute of Ophthalmology, Beijing Tongren Hospital, Capital Medical University, Beijing, China 100730

[2] Departments of Radiology and Bioengineering, University of Pittsburgh, Pittsburgh, PA, USA 15213

**Corresponding Author and Guarantor of the entire manuscript:**

Jiantao Pu, Ph.D.
Department of Radiology & Bioengineering
University of Pittsburgh
3362 Fifth Avenue
Pittsburgh, PA 15213
Tel: (412)-641-2571
Fax: (412)-641-2582
puj@upmc.edu

# ABSTRACT

**Objective:** To investigate whether and to what extent central serous chorioretinopathy (CSC) depicted on color fundus photographs can be assessed using deep learning technology.

**Methods and Materials:** We collected a total of 2,504 fundus images acquired on different subjects. We verified the CSC status of these images using their corresponding optical coherence tomography (OCT) images. A total of 1,329 images depicted CSC. These images were preprocessed and normalized. This resulting dataset was randomly split into three parts in the ratio of 8:1:1 respectively for training, validation, and testing purposes. We used the deep learning architecture termed InceptionV3 to train the classifier. We performed nonparametric receiver operating characteristic (ROC) analyses to assess the capability of the developed algorithm to identify CSC. To study the inter-reader variability and compare the performance of the computerized scheme and human experts, we asked two ophthalmologists (i.e., Rater #1 and #2) to independently review the same testing dataset in a blind manner. We assessed the performance difference between the computer algorithms and the two experts using the ROC curves, and computed their pair-wised agreements using Cohen's Kappa coefficients.

**Results:** The areas under the receiver operating characteristic (ROC) curve for the computer, rater #1, and rater #2 were 0.934 (95% CI=0.905-0.963), 0.859 (95% CI=0.809-0.908), and 0.725 (95% CI=0.662-0.788). The Kappa coefficient between the two raters was 0.48 ($p < 0.001$), while the Kappa coefficients between the computer and the two raters were 0.59 ($p < 0.001$) and 0.33 ($p < 0.05$).

**Conclusion:** Our experiments showed that the computer algorithm based on deep learning can assess CSC depicted on color fundus photographs in a relatively reliable and consistent way.

**Abstract:** central serous chorioretinopathy (CSC), fundus photography, deep learning, early screening

# I. INTRODUCTION

As the fourth most common retinopathy, central serous chorioretinopathy (CSC) is a condition where fluid accumulates under the retina and thus causes a physical detachment and vision loss [1-2]. This condition often occurs in men under a lot of stress. CSC typically appears on fundus photography as yellowish material in the superior macula, but this appearance varies significantly from case to case. It is not easy, especially for young ophthalmologists, to reliably identify CSC on fundus photography. Although color fundus photography is cost-effective and has no side effects (e.g., nausea caused by fluorescent dye), fluorescein angiography and/or optical coherence tomography (OCT) is more sensitive and is typically used to diagnose CSC in clinical practice. In the past, there have been some efforts on assessing CSC using computers based on fluorescein angiography and OCT [3-6]; however, to the best of our knowledge, there is no effort dedicated to this issue using color fundus photography. The primary reasons are the difficulty in visualizing CSC on fundus photography (e.g., no clear boundary with surrounding regions) and the variety of appearances CSC has on fundus images, as well as the confounding factors caused by the characteristics of optical imaging. We present some examinations with CSC in Figure 1 to illustrate the underlying difficulty to visually assess CSC depicted on fundus photography. These issues make it very challenging to develop efficacious algorithms to extract and quantify the underlying appearance associated with CSC. During routine screening examinations, where color fundus imaging is commonly used, it is often easy to overlook existing CSC. If a computer tool is available to automatically assess the CSC status, an ophthalmologist can perform a timely differential diagnosis to rule out some emergent medical conditions, such as retinal detachment.

Unlike traditional image processing technologies, which typically rely on explicitly extracting imaging features, the emerging CNN based deep learning technology is capable of learning imaging features or patterns in an implicit way [7-8], where tens of millions of features may be involved and analyzed. Hence, this technology may be suitable for situations where complex image texture analysis may be needed. In this study, we retrospectively collected a dataset and leveraged the deep CNN architecture to investigate whether CSC can be reliably identified using color fundus photographs. We, in particular, compared the performance of the computerized scheme and two human experts, and assessed the agreement of the raters, to study the potential of such a computer tool in practice. A detailed description of the methods and the experimental results follows.

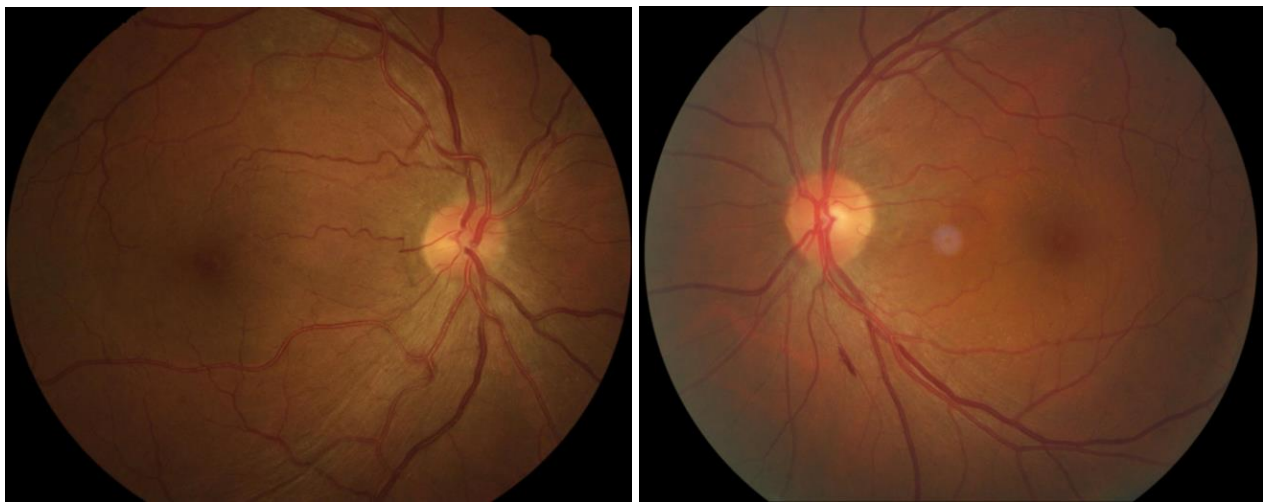

**Figure 1:** Two fundus images with verified CSC. It is not easy to visually perceive existing CSC.

## II. METHODS AND MATERIALS

### A. Data Acquisition

We collected a total of 2,504 images acquired on different subjects from the eye picture archive and communication system (Eye-PACS) at Beijing Tongren Hospital without regard for fundus camera equipment brand. All privacy information was de-identified. For each case, we have both color fundus images and OCT images. We used the OCT images to verify whether there was CSC or not. An experienced ophthalmologist was responsible for reviewing these OCT exams. Among these subjects, the ophthalmologist verified 1,329 images with CSC, and the remaining 1,175 images were negative. We summarized the demographic information in each category (e.g., gender and age), as depicted in Table 1. For machine learning, we randomly divided the dataset into three parts in the ratio of 8:1:1 respectively for training, validation, and testing purposes.

**Table 1**: Subject demographics

|  | Control group | CSC group |
|---|---|---|
| Subjects (n) | 1175 | 1329 |
| Males (%) | 859 (73.1%) | 980 (73.7) |
| Age (year) | 48.78+10.4 | 45.27+8.8 |
| Training set | 939 | 1063 |
| Validation set | 118 | 133 |
| Testing set | 118 | 133 |

B. Data Preprocessing

Given a fundus image, we processed it firstly using three steps (Figure 2). The first step was to identify the field-of-view (FOV) of the retina region depicted on a fundus image, which typically appears as a circle. For general purposes, we used an ellipse fitting operation [9] we developed to extract the FOV. The advantage of the ellipse fitting algorithm is the non-sensitivity to existing image noise / artifacts (e.g. the protruding region indicated by the arrow in Figure 2 will not be included in the region-of-interest). The second step was to normalize the image by stretching the image intensities or contrast to similar levels using the histogram equalization [10]. This approach can result in improved views and uniformed / similar appearances of the over- or under-exposed regions, which frequently happen in optical fundus images (Figure 3), thereby somewhat easing the requirement on the size of the dataset for training. The third step was to crop the FOV and resize the obtained images uniformly into 299×299 pixels in dimension for machine learning or training.

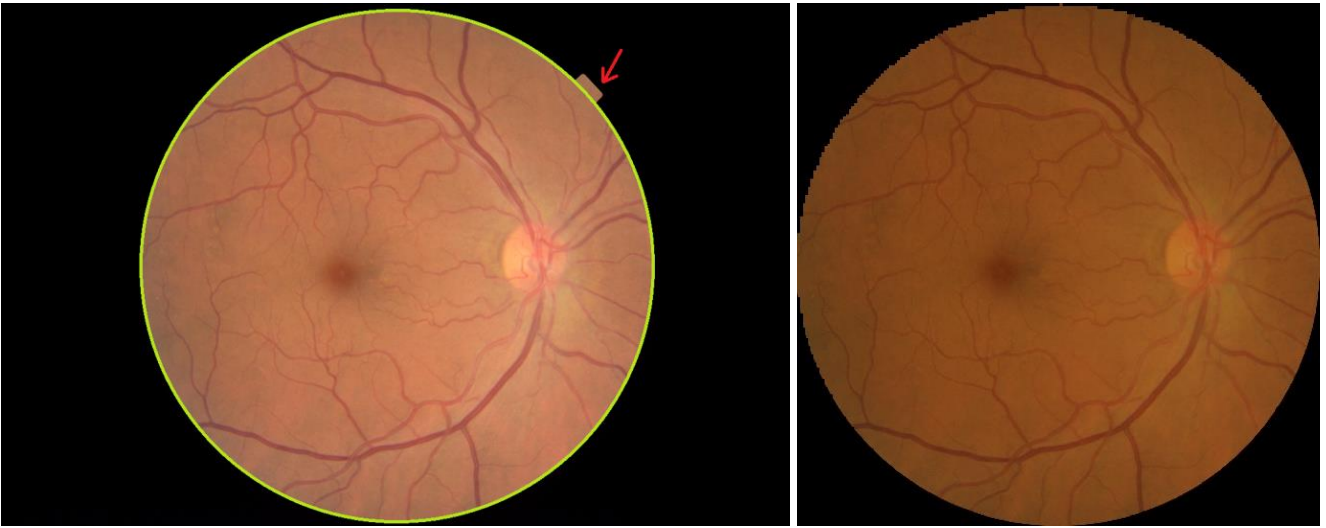

**Figure 2:** Illustration of the image preprocessing of a fundus image: (a) an ellipse fitting the field-of-view of a fundus image (i.e., the highlighted ellipse in the left); (b) the normalized, cropped, and resized image (299×299 pixels) (in the right).

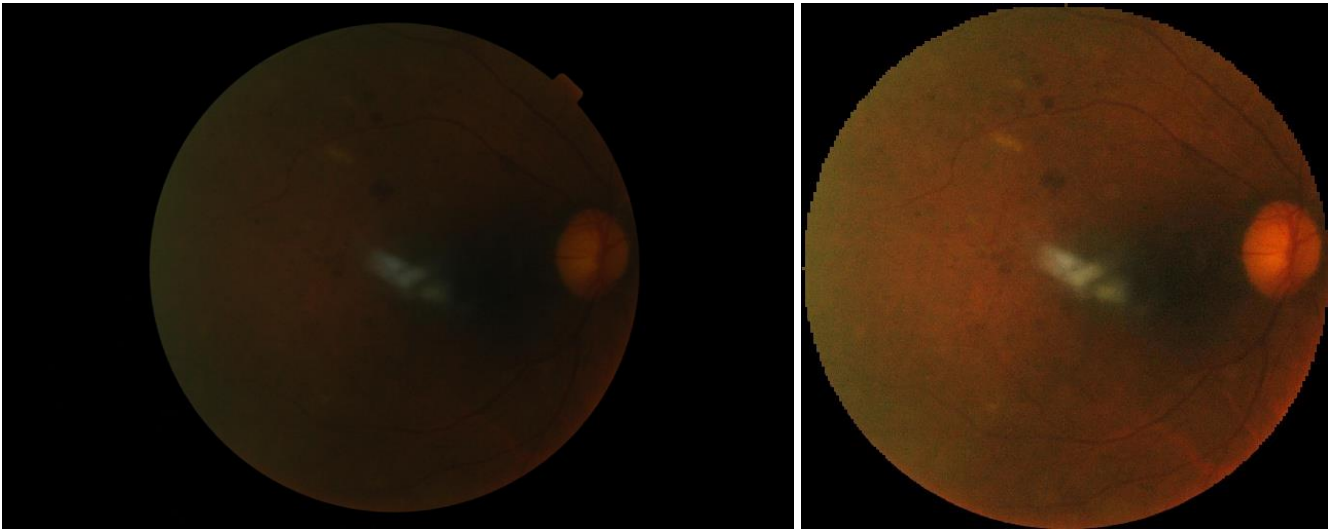

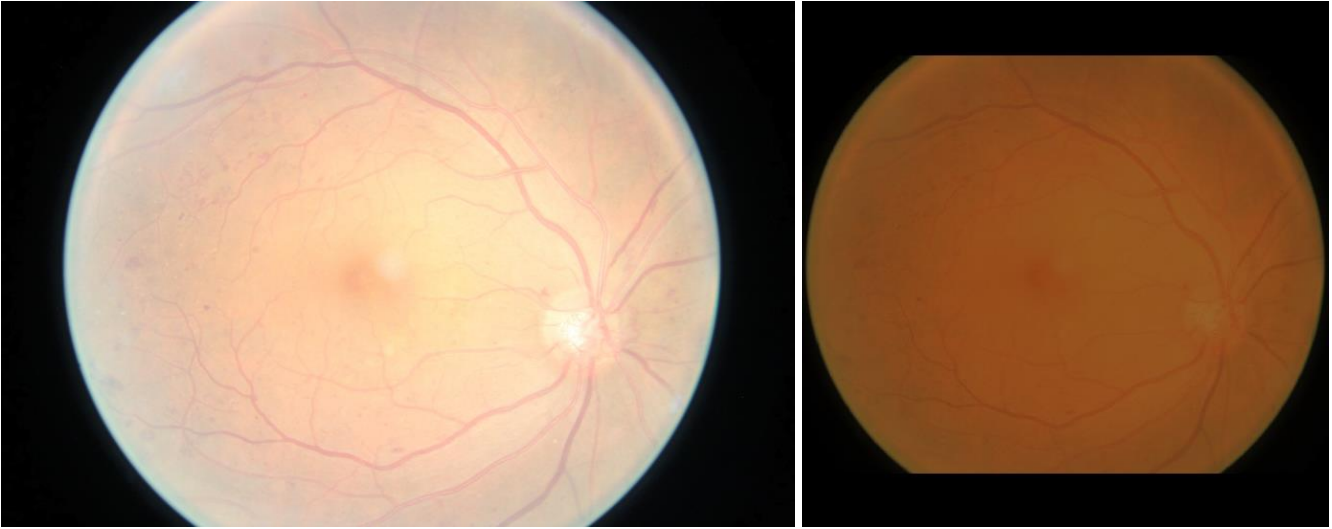

**Figure 3**: Examples demonstrating the effect of the histogram equalization. The left row showed the original fundus images, where the top row was under-exposed and the bottom row was over-exposed, and the right column showed the processed images demonstrating similar characteristics in both color and brightness.

C. **Prediction Model Implementation**

We used the well-known CNN classifier termed Inception-V3, which is a member of the GoogleNet Family [11-14]. We fine-tuned this model to take advantage of the pre-trained ImageNet weights by adding two fully-connected (FC) layers and one dropout regularization for the first FC layer. The dropout probability was 0.5, and the two FC layers had channels of 1024 and 2, respectively. The last FC layer was activated using the "softmax" function. The output of this classifier is the probability of being CSC. This classifier was implemented using the Keras deep learning library and trained using 80% of the collected cases with the Adam optimizer and 50 epochs. The training batch size was 32. The training procedure was stopped when the validation loss did not improve for 10 epochs. In particular, during the training, we augmented the data by randomly performing rotation, horizontal/vertical flip, scaling, and translation.

D. **Statistical analysis**

We assessed the performance of the developed prediction model using the independent testing dataset, among which no image was involved in the training. To evaluate the capability of the deep learning architecture in differentiating CSC, we performed nonparametric receiver operating characteristic (ROC) analysis on the testing dataset and computed the 95% confidence intervals (CI). In addition, we asked two ophthalmologists, an experienced ophthalmologist ( > 5 years) and a young ophthalmologist ( < 3 years), to independently assess CSC in the testing data and computed their agreements using the Cohen's Kappa coefficients [15]. We assessed the raters' performances using ROC analysis for comparison with the computer algorithm. In all analyses, we considered a p-value less than 0.05 statistically significant. The analyses were performed using IBM SPSS Version 24.

## III. RESULTS

The proposed model was trained using the training and the validation datasets (Table 1) on a PC equipped with a GPU (NVIDIA GeForce GTX 1070 Ti). When the training procedure was terminated, the training accuracy was 91.2% and the validation accuracy was 89.7% with a validation loss of 0.31. After applying the prediction model to the testing dataset (consisting of 118 control images and 133 CSC images), we summarized the performance of the developed model using the ROC curve in Figure 4. The AUC was 0.927 (95% CI: 0.895-0.959) and the p-value was less than 0.001. When the cutoff threshold was 0.5, the accuracy was 85.7% (215/251); specifically, 22 of 133 images in the control group was incorrectly classified as CSC, while 14 of 118 was incorrectly classified as non-CSC. We present two incorrectly classified examples in Figure 5. The assessment procedure was fully automated and it took less than 3 seconds to process a fundus image on a typical PC without GPU support.

As compared with the computerized scheme, the results and performances of the two human experts are summarized in Table 2 and Figure 4. Their classification accuracies on the same testing dataset were 83.3% and 70.9%, respectively. The Kappa coefficients indicating their agreement was 0.48 (p < 0.001), suggesting a moderate agreement. There existed a moderate agreement (Kappa coefficient:

0.58, p < 0.001) between rater #1 and the computer, and a fair agreement between rater #2 and the computer (Kappa coefficient: 0.33, p < 0.05). Both raters tended to overlook the CSC cases, namely classifying the cases as CSC negative.

**Table 2:** The summary of the correct classification results by the computer and the two raters on the testing dataset, where there were 133 CSC cases and 118 non-CSC cases.

|  | CSC | Non-CSC | Overall accuracy |
|---|---|---|---|
| rater1 | 96/133 | 113/118 | 83.3% (209/251) |
| rater2 | 62/133 | 116/118 | 70.9% (178/251) |
| computer | 111/133 | 104/118 | 85.7% (215/251) |

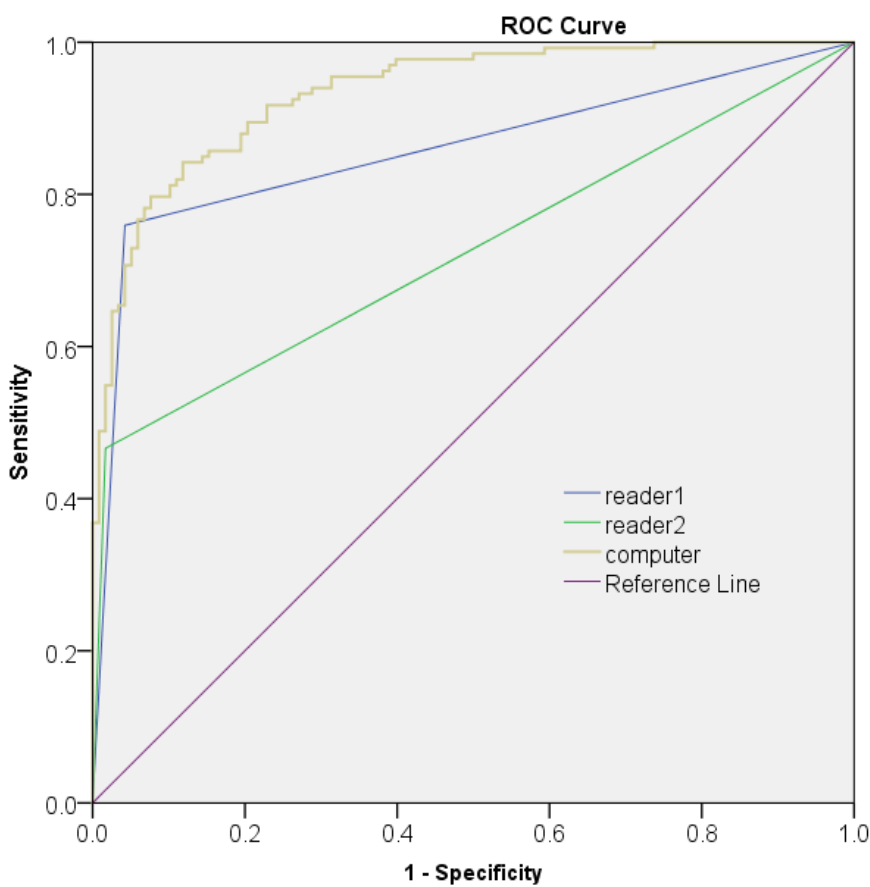

**Figure 4**: The ROC curve for identifying CSC depicted on color fundus imags. The AUCs for the computer, reader1, and reader2 were 0.934 (95% CI=0.905-0.963), 0.859 (95% CI=0.809-0.908), and 0.725 (95% CI=0.662-0.788).

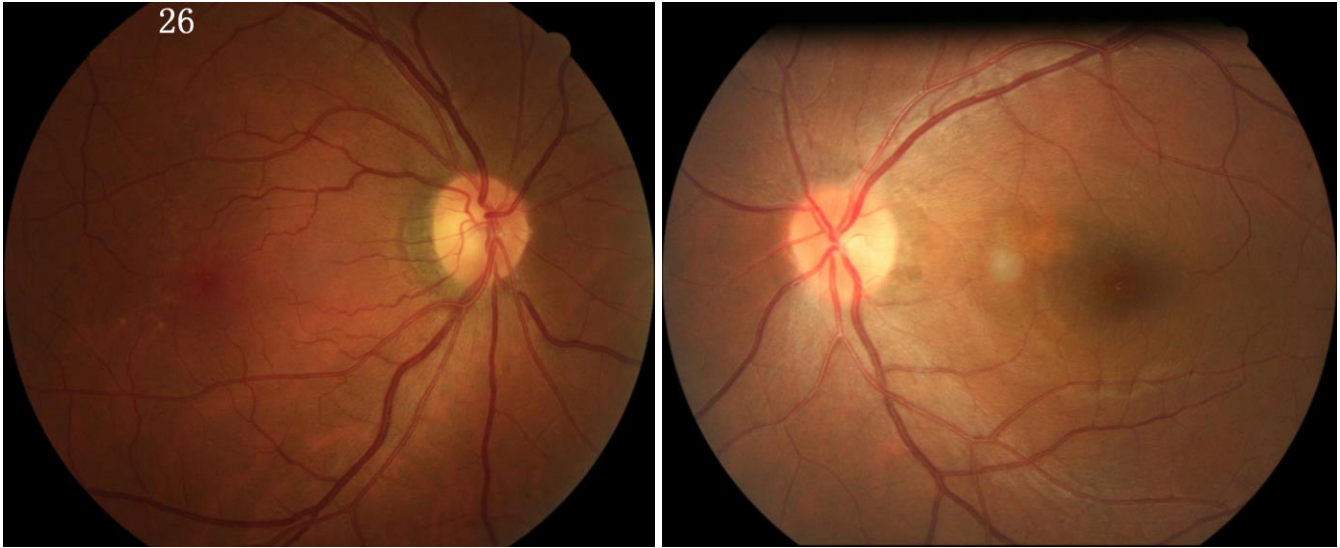

**Figure 5**: Incorrectly classified examples. The left one was mis-classified as negative and its probability of being CSC was 0.37. The right one was mis-classified as positive and its probability of being CSC was 0.68.

## IV. DISCUSSION

In this study, we proposed utilizing deep learning technology to assess CSC depicted on fundus photographs and verified its feasibility. Our statistical experiments on an independent testing dataset showed that human experts tended to have a relatively lower sensitivity as compared with the computerized scheme (Table 2). The fair agreement (Kappa coefficient: 0.33, $p < 0.05$) between the raters suggests the relatively large variability among the human experts, and experienced rater #1 performed better than unexperienced rater #2 for the given testing dataset. This suggests that unexperienced raters may overlook CSC in routine examinations. Overall, the computerized scheme achieved a better performance in classification (Figure 4), demonstrating the unique potential of the deep learning technology in screening CSC based on fundus photography.

Although there have been a number of investigations demonstrating the potential of deep learning in detecting and diagnosing retinal disease based on fundus images, such as diabetic retinopathy (DR) and diabetic macula edema (DME) [18-20], very limited work has been dedicated to identifying CSC using color fundus images. We believe that the availability of such a tool could be helpful for ophthalmologists as a second eye to timely and accurately detect CSC, given the wide usage of color fundus photography in clinical practices. In particular, it may reduce unnecessary fluorescein angiography and / or OCT examinations that may involve some type of side effect or additional cost and often not be available in developing areas. Although the developed approach achieved a reasonable performance in identifying CSC, it does not mean that the fundus image could replace the fluorescein angiography or OCT in CSC diagnosis.

There have been a number of deep learning models developed for image classification to date [16-19]. Among these, the InceptionV3 was widely used in fundus image related investigations [18-19]. This is actually the primary reason that we used the same model as these available investigations. We note that our objective in this study was to verify whether CSC can be reliably identified using fundus images and thus developing a novel deep learning model was not our emphasis. Although the developed model may not be optimal, it has demonstrated a promising performance in discriminating CSC. Also, once the prediction model is trained, assessing a fundus image becomes very efficient ($< 3$ seconds), making the screening process extremely easy and efficient.

There are some limitations with this study. First, the variety and the number of the images are limited. All these images were acquired from a single institution. In particular, for the deep learning purpose, the training and testing datasets were relatively small; however, the size of this dataset should be sufficient to demonstrate the feasibility of our conclusion, that color fundus images can be used to screen CSC in routine clinical practice. We expect that the performance may significantly improve when using a large diverse dataset for training the deep learning network and dedicating additional effort to optimize the training parameters. Second, we did not verify whether, and to what extent, existing specific diseases or image artifacts may affect CSC assessment. It is well-known that several factors could potentially affect optical fundus images, such as camera focus, light exposure, patient cooperation, and existing specific diseases (e.g., severe exudation). In appearance, these factors may cause the confusion of the CSC appearance and the resulting image artifacts. It is very challenging for the traditional feature-based approaches to handle these confounding factors and to identify the features that can reliably differentiate CSC from these factors. However, in methodology, the deep learning approach may not have these issues when big data is available for training. Finally, due to the lack of reference standard, namely that we did not discover any similar investigations that utilized fundus photography to identify CSC, we did not conduct any performance comparisons. In a general sense, the developed model demonstrated a promising and reasonable performance regardless of the above limitations and suggests the necessity of further investigations about its impact on clinical practice.

## V. CONCLUSION

We developed and validated a computerized approach based on deep learning to identify CSC depicted on color fundus images. Although we used a limited dataset for developing and validating our model, our experiments demonstrated the unique strength of the deep learning technology in this regard. Future translational effort is desirable to make such a tool available to the clinical practice for initial CSC screening and thus aid in an early and accurate diagnosis.

## ACKNOWLEDGEMENT

This work is supported in part by Grants R21-CA197493 and R01-HL096613 from National Institutes of Health.